\def\year{2021}\relax
\documentclass[letterpaper]{article} 
\usepackage{aaai21}  
\usepackage{times}  
\usepackage{helvet} 
\usepackage{courier}  
\usepackage[hyphens]{url}  
\usepackage{graphicx} 
\usepackage{natbib}  
\usepackage{caption} 
\usepackage[usenames,dvipsnames]{xcolor}
\usepackage{graphicx, epsfig, cite, acronym, amssymb, algorithm, algorithmic}
\usepackage[normalem]{ulem}
\newcommand{\rev}[1]{\textcolor{blue}{\uline{#1}}}  

\title{Two ways towards combining Sequential Neural Network and Statistical Methods to Improve the Prediction of Time Series}
\author{

    Jingwei Li\\
}

\affiliations{
    Department of Electrical and Computer Engineering, Stony Brook University, NY, 11790

}

\begin{document}

\maketitle


\begin{abstract}

    Statistic modeling and data-driven learning are the two vital fields that attract many attentions. Statistic models intend to capture and interpret the relationships among variables, while data-based learning attempt to extract information directly from the data without pre-processing through complex models. Given the extensive studies in both fields, a subtle issue is how to properly integrate data based methods with existing knowledge or models. In this paper, based on the time series data, we propose two different directions to integrate the two, a decomposition-based method and a method exploiting the statistic extraction of data features. The first one decomposes the data into linear stable, nonlinear stable and unstable parts, where suitable statistical models are used for the linear stable and nonlinear stable parts while the appropriate machine learning tools are used for the unstable parts. The second one applies statistic models to extract statistics features of data and feed them as additional inputs into the machine learning platform for training. The most critical and challenging thing is how to determine and extract the valuable information from mathematical or statistical models to boost the performance of machine learning algorithms. We evaluate the proposal
    using  time series data with varying degrees of stability.  Performance results show that both methods can outperform existing schemes that use models and learning separately, and the improvements can be over 60\%. Both our proposed methods are promising in bridging the gap between model-based and data-driven schemes and integrating the two to provide an overall higher learning performance.

\end{abstract}
\section{Introduction}

In recent years, we are seeing a rapid growth of data from the increasing deployment of sensors and sources of various types. The availability of massive data calls for new and more advanced schemes to analyze the data, and apply the knowledge to  enable more intelligent and powerful applications.  Data-based (model-less) applications attempt to extract information directly from the data without preprocessing through complex models. Machine learning systems are becoming increasingly employed in complex  settings such as medicine (e.g. radiology, drug development), financial technology (e.g. stock price prediction, digital financial advisor), and even in law enforcement (e.g. case summarization, litigation prediction). Despite this increased utilization, there is still a lack of sufficient techniques to explain and interpret the decisions of these deep learning algorithms.

The convergence of data and model has been reflected from at least two aspects. First, massive data have transformed the way how system states can be estimated and predicted, which is essential for reliable system operations and decision making. Second, data also provide a venue to critically examine and refine traditional “model-based” methods for many applications. These have led to the “data-driven” methods that have greatly empowered the applications. However, a subtle issue is how to properly integrate data-driven methods with existing knowledge or models. After all, many application fields have been extensively studied for the last almost a century, and has well-established theories and models. Data, big or small, can better reflect the knowledge or help discover new knowledge, but are not expected to fundamentally change the established knowledge. Therefore,maximizing the potential of data-driven methods while at the same time respecting basic theories and knowledge to enable new and powerful applications is currently a grand challenge.

The introduction of useful new information may be helpful to  improve the performance of both the model-based and data-based learning. For example,  to predict the daily "opening price" of a stock\cite{sureshkumar2012performance}\cite{ariyo2014stock}\cite{di2016artificial}, it would be helpful to introduce additional information such as the daily "closing price", "highest price", "lowest price" and "trading volume". However, a more valuable research is how to improve the learning performance by making the best use of models and existing data without requiring new information to add in the cost and delay. Given a model already established to describe an application and the large amount of data that reflect practical states, the question is if these two can be integrated and how they can work together to provide an overall new information paradigm. 

To fundamentally advance the learning performance while being able to control and interpret the results,  we propose two methods to explore two directions of effectively integrating the model and data. We use the learning of sequential data distribution as an example to explain our design principles, while our methods can be extended to more general learning scenarios. In the first method,  data are decomposed into three parts, {\em linear stable, nonlinear stable and unstable.} The first two parts are represented with mathematical models, while the  unstable part is learnt  through pure data-based methods. Rather than purely relying on mathematical model or machine learning, we expect the model can better represent and interpret the stable data with prior knowledge of the data features, while the pure data-driven learning can help to capture the uncertainty and dynamics of data in a practical scenario.  The second method exploits data-driven learning to more flexibly capture the data distribution, while taking advantage of mathematical models to extract statistics of data and feed them as new features into the machine learning platform. The focus is on the finding of proper statistics and the way to integrate model-based and pure data-driven learning. It does not require capturing new data to obtain the additional features. 

In the rest of the paper, we first provide background knowledge on statistical models and machine learning methods commonly used to represent time series. We then present the details of our two methods of integrating mathematical models with the data-driven learning scheme. We compare the performance of our schemes with those purely using models or relying on data, and the results show that our schemes are very effective and achieve the best performance in all data sets.  

\section{Preliminaries}

Various types of statistical models\cite{aamir2016modelling}\cite{gupta2019very}\cite{shbier2017swgarch}\cite{zhu2020high} have been developed in the literature to represent the time series data for different applications. In recent years, increasing attentions are drawn to apply neural networks and deep learning\cite{kashiparekh2019convtimenet}\cite{sangiorgio2020robustness}\cite{hewamalage2020recurrent} \cite{zhu2021news}\cite{zhu2020adaptive} represent complex data distributions. Before presenting our proposed methods, we review some basic statistical models commonly used to represent financial data, and basic neural network structure applied to track time series data. 

\subsection{Statistic Models}
Compared to methods using the neural networks, statistic models can better explain the relationship between data and reveal the mechanism of change among data. In addition, model-based methods have good analytical properties, and can better calculate and theoretically prove the existence and convergence of errors. Taking daily financial stock data as an example, the data constantly change over time. Parameters, such as daily open price, close price, maximum price and minimum price, always change but have some relationships with the data of previous days. For example, the daily close price in day $t$ may be related with those in day $t-1, t-2…t-10$. Generally, the influences from closer days are bigger.  Time series models intend to find and capture the strong relationship among the data in day $t$ and the data in previous days. A few models are commonly used to represent financial data. 
\paragraph{A. ARMA Model}
\vspace{-0.05in}
Auto-Regressive Moving Average (ARMA)\cite{brockwell2002introduction} is a general model used to forecast a stationary time series. ARMA ($p,q$) combines the Auto-Regressive (AR ($p$)) process and the Moving Average (MA($q$)) process\rev{.} AR($p$) represents a stationary time series
\begin{eqnarray}
				y_t = \mu + \alpha_1{y_{t-1}}+\alpha_2{y_{t-2}}+...+\alpha_p{y_{t-p}}+e_t,
\end{eqnarray}
where $\mu$ is a constant, $\alpha_1,\alpha_2,...\alpha_p$ are the auto-correlation coefficients at lags 1,2…$p$, and the residual $e_t$ is often assumed to be Gaussian white noise with the mean zero and the variance $\sigma_t$.  To improve the accuracy of time series prediction,  MA ($q$) model takes into account  the historical impact of white noise to predict the sequential value: 


\begin{eqnarray}
	y_t = \mu + e_t+\theta_1{e_{t-1}}+\theta_2{e_{t-2}}+....+\theta_q{e_{t-q}},
\end{eqnarray}
where $\mu$ is the expectation of $y_t$ (usually assumed equal to zero), $\theta$ terms are the weights for prior stochastic term in time series. $e_t$ is often assumed to follow Gaussian white noises with mean zero and variance $\sigma_t$. 
Integrating AR($p$) and MA($q$), ARMA($p$,$q$) model is expressed as
\begin{eqnarray}
   y_t = \mu + \sum_{i=1}^{p} {\alpha_i{y_{t-i}}} + e_t+ \sum_{j=1}^{q} {\theta_j{e_{t-j}}}.
\end{eqnarray}

\paragraph{B. ARIMA Model}
\vspace{-0.05in}
 To model the non-stationary time series data,  Auto-Regressive Integrated Moving Average (ARIMA)\cite{weisang2008vagaries} is used to generalize ARMA model with ARIMA ($p$,$d$,$q$), where  a difference factor or some nonlinear transformation (including power transformation and logarithmic transformation) is introduced. $d$ is the number of difference items needed to convert a non-stationary time series into a stationary one. An  ARIMA($p$,$d$,$q$) time series will follow the ARMA($p$,$q$) model after $d$ times of difference. For example, if a time series $y_t$ follows the ARIMA($p$,$d$,$q$) model, then $\triangle^d$$y_t$ follows the ARMA($p$,$q$) model, where $\triangle^d$$y_t$ is the sequence of $y_t$ after $d$ times differences. 
 
\paragraph{C. ARCH Model}
\vspace{-0.05in}
In the ARIMA model, we assume the errors follow the homogeneous Gaussian distribution, while Auto-Regressive Conditional Heteroscedasticity (ARCH) model\cite{engle1995arch} is used when errors do not follow the same distribution but change over time. The variance of the errors at time $t$ is affected by the errors before $t$, and  is thus considered to be conditional. The ARCH($P$) model has the following structure:

\begin{eqnarray}
&& y_t  =  \mu_t +  u_t,   \quad u_t = z_t{\sqrt{h_t}},\quad z_t\sim N(0,1), \nonumber\\
&& h_t = w+\sum_{j=1}^{P} {\lambda_j{u_{t-j}^2}}. 
\end{eqnarray}



where the constant $\mu_t$ is usually assumed to equal the expectation of the time series. The random error $u_t$ is a function of $z_t$ (which follows a normal distribution)  and the GARCH term $h_t$ (which represents a conditional variance).  $\lambda_j$ is a weight. 





\paragraph{D. GARCH Model and ARMA-GARCH Model}
\vspace{-0.05in}
The conditional variance functions of some residual time series often have the feature of auto-regression, where the current conditional variance is also affected by the previous  variance values. As a generation of the ARCH model, the GARCH($P,Q$) model\cite{francq2019garch} follows

\begin{eqnarray}
 &&  y_t = \mu_t + u_t ,   u_t = z_t{\sqrt{h_t}},\quad z_t\sim N(0,1),\nonumber\\
 && h_t = w+\sum_{j=1}^{P} {\lambda_j{u_{t-j}^2}}+\sum_{i=1}^{Q}{\beta_i{h_{t-i}}}.
\end{eqnarray}

where $h_t$ is the conditional variance. GARCH model can be integrated with ARMA model as
\begin{eqnarray}
&& y_t = \mu_t + u_t,  u_t = \sum_{i=1}^{p} {\alpha_i{u_{t-i}}} + e_t+ \sum_{j=1}^{q} {\theta_j{e_{t-j}}},\nonumber\\
&& e_t = z_t{\sqrt{h_t}},\quad z_t\sim N(0,1),\nonumber\\
&& h_t = w+\sum_{j=1}^{P} {\lambda_j{e_{t-j}^2}}+\sum_{i=1}^{Q}{\beta_i{h_{t-i}}}.
\end{eqnarray}

to form an ARMA($p$,$q$)-GARCH($P$,$Q$) model. If $y_t$ is not stationary, it can be replaced with a $d$ times difference series $\triangle^d$$y_t$  to form the ARIMA($p$,$d$,$q$)-GARCH($P$,$Q$) model.

\subsection{Data-driven Learning}

Rather than following a pre-established model, modern machine learning and deep learning methods derive the knowledge directly from data  
without assuming their distribution format, and can well follow nonlinear data. However, they often suffer from large computational cost, weak interpretability,  and bias when data are unbalanced. In the case of time series, 
sequential neural networks, such as Recurrent Neural Network (RNN), Long Short-Term Memory (LSTM) and Gated Recurrent Units (GRU), are often applied.

\paragraph{A. Recurrent Neural Network}

RNN \cite{manaswi2018rnn}\cite{tokgoz2018rnn}\cite{poulos2017rnn}includes a neural network to provide a very straight-forward but effective way of handling time series or other sequential data. RNN is recurrent, where the same function is performed for each time stage and the output of the previous time stage is the input of the next stage. RNN can model a sequence of data so that each sample  depends on previous ones. Backward Propagation Algorithm is often used to train RNN. 



\paragraph{B. Long Short-Term Memory}
As the back-propagated error  either explodes or vanishes during the propagation in the training process, RNN cannot process very long time sequence.  To alleviate the problem, 
LSTM \cite{song2020time}\cite{sagheer2019time}
is introduced. It has two transmission states, the  cell state $c_t$  and the hidden state $h^*_t$. $c_t$  changes slowly, while $h^*_t$ changes faster. 
An LSTM time stage consists of many recurrently
memory cells. Each cell contains three multiplicative gate units, named input gate, forget gate and output gate.

\section{Modeling Time Series Data with Mathematical Models and Data-driven Learning}



 
 Generally, statistical models can well represent stable data with slow variations, but are not good at tracking dynamic data with fast changes. 
On the other hand,  without assuming the data format, learning directly from data can more flexible represent data. However, if there is no knowledge on the data,  it may suffer from long training time. It is also hard to control the learning process and explain the results.
 In order to more accurately model the (time series) data while reducing the complexity and increasing the interpretability, we explore the use of both  model-driven  and data-driven approaches.


We first introduce a decomposition-based method to exploit both types of techniques.  We decompose each data item into multiple parts, and for each part, we choose a specific model or learning format that can best represent its data features. Given the tradeoff between methods that are purely based on models and purely driven by data, if the statistics of data are known, it may help to 
better understand the data characteristics. Therefore, as a second method, we propose to extract the statistics of data and incorporate them into the pure data-driven learning framework to improve the performance. 

\subsection{Method 1: Modeling Time Series Data with Decomposition}

Time series data can be divided into a stable part and an unstable part. Generally, the stable part can be more easily modeled, while it is often hard to  accurately model the unstable part. Further, the stable part can be categorized  into two types, linearly stable  and non-linearly stable. From the model features we introduce earlier, we could choose to model the linearly stable part of the data with ARIMA and the non-linearly stable part with GARCH. In order to capture all possible features in practical data, we can represent the data with  ARIMA-GARCH and Machine Learning techniques (ML) together, so both stable part and and unstable part of the data can be tracked. 

A few ML approaches have been introduced to represent time series data. 
LSTM is a  popular ML method to deal with time series with a recurrent neural network. We use  LSTM to learn the unstable part,
and represent the overall data with $ARIMA(p,d,q)-GARCH(P,Q)-LSTM$. It is not difficult to see that this complete data representation can reduce to $ARIMA(p,d,q)-LSTM$ when there are only linear stable part and unstable part in the time series, that is ($P=0,Q=0$). The complete $ARIMA-GARCH-LSTM$ model  will reduce to the $GARCH($P$,$Q$)-LSTM$ model if there is not linearly stable part in the time series, that is ($p=0,q=0$). Next we we introduce the detailed formats of these three representations.

\paragraph{A. Complete data representation with ARIMA-GARCH-LSTM}

As a general format, a time series $y_t$ can be represented as:
\begin{eqnarray}
     y_t = S_t + N_t = LS_t + NS_t + N_t
\end{eqnarray}
where $S_t$ is the  stable part, $N_t$ is the unstable part, $LS_t$ is the linearly stable part and $NS_t$ is the non-linearly stable part.  $S_t$ can be modeled by ARIMA-GARCH and $N_t$ can be learnt through $LSTM$. In this case, $y_t$ can be expressed more completely as:
\begin{eqnarray}
&& y_t = \mu_0  + u_t + \widehat{LSTM}, \nonumber \\
&& u_t = \sum_{i=1}^{p} {\alpha_i{u_{t-i}}} + e_t+ \sum_{j=1}^{q} {\theta_j{e_{t-j}}},\nonumber\\
&& e_t = z_t{\sqrt{h_t}}, \quad  z_t\sim N(0,1), \nonumber \\
&& h_t = w+\sum_{j=1}^{P}{\lambda_j{e_{t-j}^2}}+\sum_{i=1}^{Q}{\beta_i{h_{t-i}^2}}
\end{eqnarray}

where $y_t$ is a stationary time series, and $\mu_0$ is a constant. The stable part $u_t$ follows the $ARIMA-GARCH$ model and the unstable part  is learnt through $LSTM$. The random error $e_t$ is a function of $z_t$, which follows the normal distribution,  and $h_t$ is the GARCH term  (i.e. conditional variance). The parameters $\alpha_i$, $\theta_j$, $\beta_i$ and $\lambda_j$ are the weights.

\paragraph{B. ARIMA-LSTM}

If a time series just includes a linearly stable part and an unstable part, the time series can be represented as
\begin{eqnarray}
     y_t = LS_t +  N_t,
\end{eqnarray}
where $LS_t$ can be modeled by ARIMA model. The equation above can be expressed as
\begin{eqnarray}
&& y_t = \mu_0 + u_t + \widehat{LSTM} \nonumber\\
&& u_t = \sum_{i=1}^{p} {\alpha_i{u_{t-i}}} + e_t+ \sum_{j=1}^{q} {\theta_j{e_{t-j}}}
\end{eqnarray}
\paragraph{C. GARCH-LSTM}

If a time series just includes a non-linearly stable part and an unstable part, the time series can be represented as
\begin{eqnarray}
     y_t = NS_t +  N_t
\end{eqnarray}
where $NS_t$ can be modeled by the GARCH model. The equation above can be expressed as
\begin{eqnarray}
&& y_t = \mu_0 + u_t + \widehat{LSTM}, u_t = z_t{\sqrt{h_t}},\quad z_t\sim N(0,1),\nonumber \\
&& h_t = w+\sum_{j=1}^{P} {\lambda_j{u_{t-j}^2}}+\sum_{i=1}^{Q}{\beta_i{h_{t-i}^2}} 
\end{eqnarray}


\subsection{Method 2: Sequential Neural Network with Statistic Extraction}
 Good statistical values of a time series can help to better understand and capture the characteristics  of data. The average, median, mode and quantile are some example statistics parameters that can reflect the distribution of a data set and are used the most. 
 However, if these statistics are only drawn from existing data and set to fixed values, they can not well reflect the time varying features of time series data. If the statistics themselves can be modeled as  time series,  they may be fed as additional input variables to help better train the sequential neural network.
 
 Apart from variables commonly used as inputs to the sequential neural network, we add in the statistics extracted from the data series as additional inputs. The GARCH term $h_t$ is a good candidate to choose. As a conditional variance, it can reflect the volatility of the sequence at different time and is essential for describing and predicting the changes of sequence. To explain our design principles, we use LSTM as an example neural network and the GARCH term $h_t$ as the example statistics. 

  We use the past values of both $y_t$ and $h_t$ to train LSTM and predict the future values of $y_t$.  That is, besides past values of $y_t$ (i.e., ${y_{t-1}, y_{t-2},...,y_{t-k}}$), we add past values of the GARCH term $h_t$ (i.e., ${h_{t-1}, h_{t-2},...,h_{t-k}}$) as another input: 
 
 \begin{eqnarray}
    {y_{t}} = LSTM(y_{t-1},y_{t-2},...,y_{t-k};h_{t-1},h_{t-2},...h_{t-k}) + \varepsilon_t
\end{eqnarray}
where $\varepsilon_t$ is the random error. For the convenience of expression, we call the model above as LSTM-GARCH model. 

\section{Performance Evaluation}
We compare the performance of our model with several reference models on the prediction of stock  data. 

\subsection{Experimental Setup}
\subsubsection{Data sets}
As typical time series data, stock prices are commonly used in data analyses. We choose three data sets on "Open Price" of different stocks in our performance studies, with each  set containing the data of one year. All the data set can be obtained from  \url{https://eoddata.com/stockquote/NASDAQ/AMZN.htm.}   
The first data set is the daily open stock price of ”AMZON(AMZN)” from 01-02-2019 to 12-18-2019. The second data set is the daily open stock price for "Fidelity National Information Services (FIS)" from 01-02-2019 to 12-18-2019. The third one contains the "Federal Home Loan Mortgage Corp(FMCC)" Company from 01-02-2018 to 12-04-2018. We divide each data set into two parts, $91\%$ data are put in the training set and the remaining $9\%$ is used as the test set. Each training set has 214 samples and each test set has 21 samples. In order to further investigate the performance of our proposed methods, we  divide the test data into three subgroups: test set 1 with one week's data from day 1 to day 7, test set 2 with two weeks' data from day 1 to day 14 and test set 3 with three weeks' data from day 1 to day 21. The first data sample in the test set is considered as the data from Day 1. The first data set is represented as $y_t$ and used as the primary example to show the performance of our methods. 

\subsubsection{Preparation for ARIMA-GARCH Model } Statistic models are often impacted by the stationary level of the data. We use ARIMA-GARCH model in our method 1. As a first step, we check if data are stationary.
 A time series {$x_t$} is strictly stationary if $f(x_1,...,x_n) = f(x_{1+h},...,x_{n+h})$, where $n$ and $h$ are positive integer numbers and $f(\bullet)$ denotes a function. ARMA-GARCH can only deal with stationary data, while ARIMA-GARCH can transfer  non-stationary data into stationary ones.

In statistics, Augmented  Dickey-Fuller Test(ADF) test is the most popular method to help check if a time series is stationary, and the null hypothesis of ADF test indicates that the time series is not stationary.  The p-values obtained by the test is used to decide whether we should reject the null hypothesis, so the times series is considered to be stationary. If p-value is bigger than 0.05, we will accept the null hypothesis; otherwise, we will reject the null hypothesis. In our ADF test of data, the p-value is 0.6135,  so we accept the null hypothesis and conclude that the data set is not stationary. Then we transfer the data set $y_t$ to the difference form $\triangle$$y_t$. From the ADF test of $\triangle$$y_t$,  we get the p-value 0.01, so we can reject the null hypothesis and conclude that $\triangle$$y_t$ is stationary. Although $y_t$ is not stationary, its 1 time difference is, so the parameter $d$ in the ARIMA(p,d,q)-GARCH(P,Q) model is 1. Next we introduce our procedures in constructing the ARIMA-GARCH model for $y_t$. 

\subsubsection{ARIMA(p,d,q)-GARCH(P,Q) Model}
 We use R Studio to build up the ARIMA-GARCH model for the data. Using the "auto.arima" function from 'tseries' library in the studio and the AIC criterion, we determine that $p=5, d=1, q=2$ for ARIMA (p,d,q). We further check whether  the variance of the residuals are homogeneous to determine if there exists an arch effect. If the variance is not homogeneous,  we need to build an GARCH model to capture the data dynamics. For the dataset ${y_t}$, we form the ARIMA(5,1,2)-GARCH(1,1) model based on the AIC criterion.



To use the model for prediction, we first build the ARIMA(p,d,q)-GARCH(P,Q)  model using the training data. More specifically,  we first use the training data ${y_1,y_2,...,y_m}$ to predict $y_{m+1}$. Then the true value of $y_{m+1}$ will be added and we use ${y_1,y_2,...,y_m,y_{m+1}}$ to predict $y_{m+2}$, 
where the coefficients $\alpha_i$ and $\theta_j$ of ARIMA(p,d,q)-GARCH(P,Q) are updated but not the model structure $(p,d,q)$ and $(P,Q)$.

\subsubsection{Setting of the decomposition method}




We use LSTM  as  the baseline method to directly learn the data distribution and make one step forward prediction.  The "Keras" version '2.3.1', a high-level neural network API written in Python, is used to build up LSTM. 
We choose " Mean squared Error" as the loss function and  $tanh$ as activation function.

In our proposed ARIMA-GARCH-LSTM, we apply ARIMA-GARCH to model the stable data components, and LSTM  to learn the distribution of the residual data ${r_t}$.  
To predict ${y_{t+1}}$, the predicted ${r_{t+1}}$ is added to the part of ${y_{t+1}}$ that is predicted with ARIMA-GARCH.

\subsubsection{Setting of the method based on statistics extraction} The GARCH term--$h_t$, obtained from the equation 8, is a statistics that can reflect the volatility of a time sequence at the time $t$. In our proposed statistics extraction method, to help capture the dynamic of $y_t$, the ARIMA-GARCH model is first built up for the data to obtain $h_t$, which is then used as a new feature to input into the LSTM framework with $y_t$.


To evaluate the effectiveness of our proposed method, we adopt four common metrics:  Mean Squared Error(MSE), Root Mean Squared Error (RMSE), Mean Absolute Error (MAE),and Mean Absolute Percentage Error (MAPE).  For all the metrics, the lower the value, the better the performance of the corresponding models. Using $y$ to denote the ground truth and $\hat{y}$ to denote the corresponding forecast,  the four metrics based on $n$ predictions are defined as follows: 
\begin{eqnarray}
&&  MSE=\frac{1}{n}\sum_{i=1}^{n}(\hat{y_i}-y_i)^2,  MAE = \frac{1}{n}\sum_{i=1}^{n}{\mid\hat{y_i}-y_t\mid}, \nonumber\\
&& RMSE = \sqrt{\frac{1}{n}\sum_{i=1}^{n}({\hat{y_i}-y_i})^2} \nonumber\\
&& MAPE = \frac{100\%}{n}\sum_{i=1}^{n} \mid\frac{\hat{y_i}-y_i}{y_i}\mid
\end{eqnarray}
\subsection{Result Analysis}


\subsubsection{Results of Decomposition Method}
Table \ref{tab1} shows that ARIMA-GARCH-LSTM  outperforms existing purely model-based scheme  ARIMA-GARCH and purely data-driven scheme LSTM  in all metrics. Taking results in data set 1 as an example, MSE, RMSE, MAE and MAPS of our proposed model improve $66.56\%$, $42.17\%$, $50.57\%$ and $50.35\%$ 
respectively compared with LSTM and $30.97\%$, $16.92\%$, $20.80\%$ and $20.71\%$ respectively compared with ARIMA-GARCH.    The results demonstrate that our decomposition-based method can effectively integrate the statistic model with the purely data-driven scheme to largely improve learning performance.



Comparing the results across the subsets of the test data, the performance improvement of the test set 1 is the biggest.   For the test set 3, the improvements of MSE, RMSE, MAE and MAPE are $40.19\%$, $22.66\%$, $28.77\%$ and $28.42\%$ respectively compared with LSTM and $8.43\%$, $4.31\%$, $11.30\%$ and $11.22\%$ respectively compared with ARIMA-GARCH model. This is because we build our model on the base of  ARIMA(p,d,q)-GARCH(P,Q). From the assumption of ARIMA-GARCH model  equation (6) and our preliminary study, its residual has a zero mean. The predicted value  $\hat{y_t}$ is equal to the summation of predicted ARIMA-GARCH value $\hat{AG}$ and the predicted residual $\hat{r_t}$, i.e., $\hat{y_t}$ = $\hat{AG}$ + $\hat{r_t}$. 
Accordingly, we have $E(\hat{y_t})$ = $E(\hat{AG})$ + $E(\hat{r_t})$ = $E(\hat{AG})$. If the test set has a long time duration, the overall performance of ARIMA-GARCH-LSTM will approach that of ARIMA-GARCH.

\begin{figure}
  \vspace{-10pt}
  \begin{center}
\includegraphics[width=2.5 in]{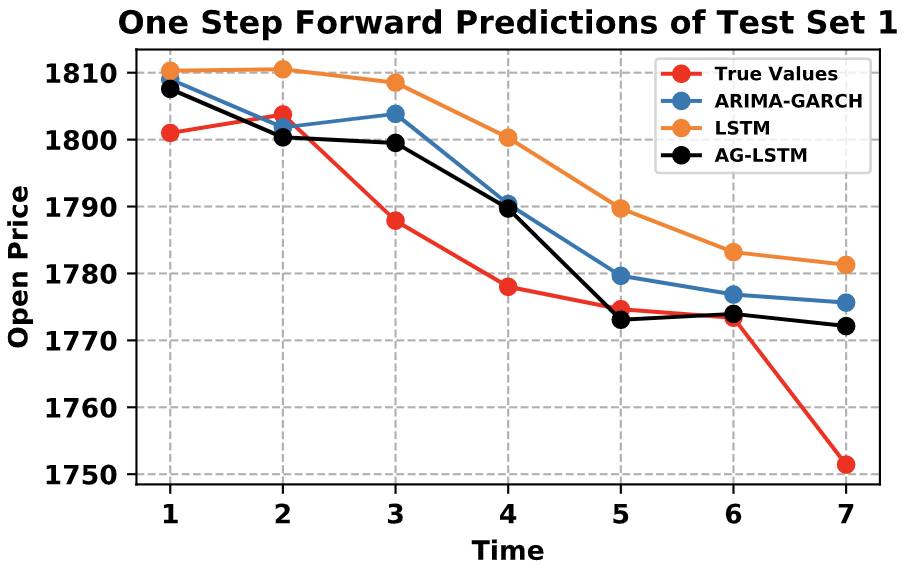}
  \end{center}
  \vspace{-10pt}
  \caption{\footnotesize Predictions of AG-LSTM and its baselines for test set 1}
\label{AGL7}
  \vspace{-5pt}
\end{figure}

\begin{figure}
  \vspace{-5pt}
  \begin{center}
\includegraphics[width=2.5 in]{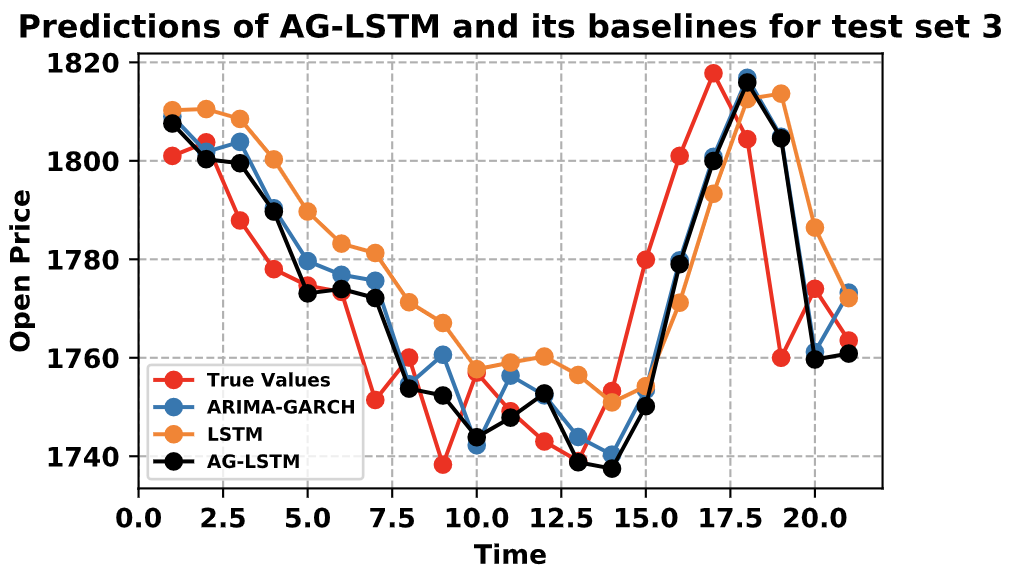}
  \end{center}
  \vspace{-10pt}
  \caption{\footnotesize Predictions of AG-LSTM and its baselines for test set 3}
\label{AGL28}
  \vspace{-5pt}
\end{figure}

Figures \ref{AGL7} and \ref{AGL28}  show the prediction over time for the data in the test sets 1 and  3 respectively. 
ARIMA-GARCH-LSTM is shown to be the closest to the ground-true curve in all results, and the accuracy is higher in the prediction of data in the testset 1.
Moreover, we observe a first-order-lag in all curves, which means the prediction on the day $t$ is more close to the result in the day $t-1$. Taking figure ~\ref{AGL7} as an example, if we move the true data from 1-6 to the position 2-7, the results from all methods have the same trend of change. The lag problem is  common  in the time series analysis, as the prediction results of all methods have a tendency to approach those of the previous step.

\begin{table}[]
\footnotesize
\begin{tabular}{|c|c|c|c|c|c|}
\hline
 &  & MSE & RMSE & MAE & MAPE \\ \hline
 & AG & 156.93 & 12.53 & 10.13 & 0.57 \\ \cline{2-6} 
Test Set 1 & LSTM & 323.92 & 18.00 & 16.23 & 0.90 \\ \cline{2-6} 
 & \begin{tabular}[c]{@{}c@{}}AG-\\ LSTM\end{tabular} & \textbf{108.32} & \textbf{10.41} & \textbf{8.02} & \textbf{0.45} \\ \hline
 & AG & 155.33 & 12.46 & 10.56 & 0.60 \\ \cline{2-6} 
Test Set 2 & LSTM & 280.57 & 16.75 & 14.38 & 0.81 \\ \cline{2-6} 
 & \begin{tabular}[c]{@{}c@{}}AG-\\ LSTM\end{tabular} & \textbf{108.06} & \textbf{10.40} & \textbf{8.34} & \textbf{0.47} \\ \hline
  & AG & 287.66 & 16.96 & 13.92 & 0.78 \\ \cline{2-6} 
Test Set 3 & LSTM & 440.38 & 20.99 & 17.34 & 0.97 \\ \cline{2-6} 
 & \begin{tabular}[c]{@{}c@{}}AG-\\ LSTM\end{tabular} & \textbf{263.40} & \textbf{16.23} & \textbf{12.35} & \textbf{0.70} \\ \hline
\end{tabular}
\vspace{-5pt}
\caption{\label{tab1} Comparison of the decomposition-based ARIMA-GARCH-LSTM with reference schemes on different test sets}
\end{table}

\begin{figure}
  \vspace{-5pt}
  \begin{center}
\includegraphics[width=2.5 in]{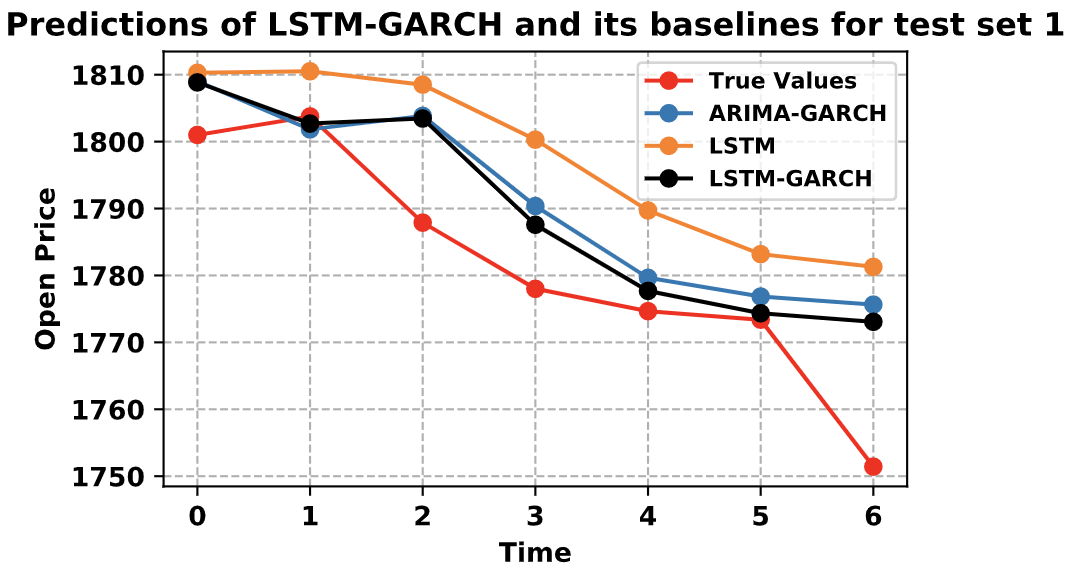}
  \end{center}
  \vspace{-10pt} 
  \caption{\footnotesize Predictions of LSTM-GARCH and its baselines for test set 1}
\label{LG7}
  \vspace{-10pt}
\end{figure}

\begin{figure}
  \vspace{-10pt}
  \begin{center}
\includegraphics[width=2.5 in]{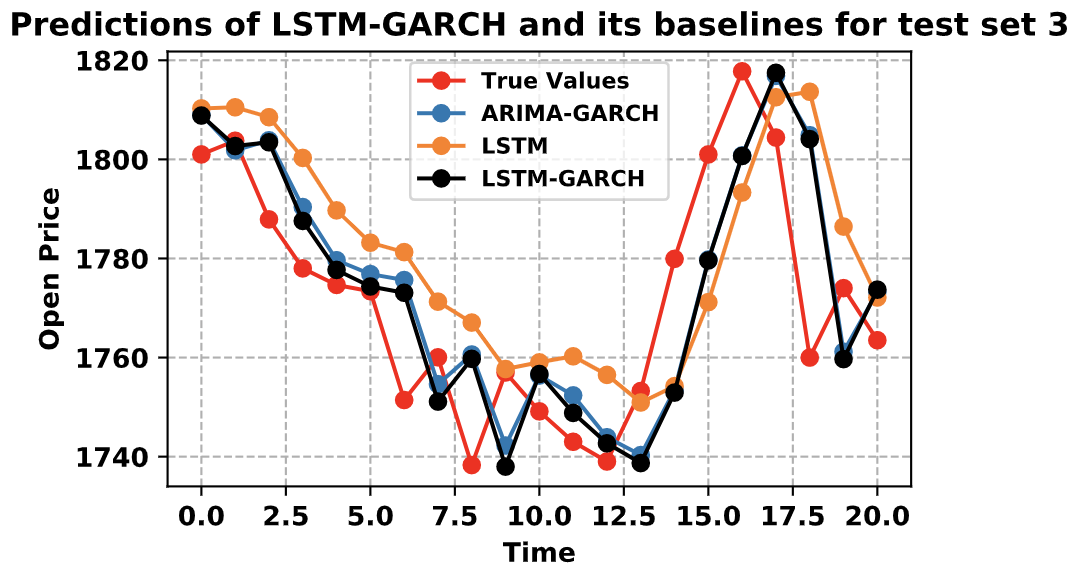}
  \end{center}
  \vspace{-10pt}
  \caption{\footnotesize Predictions of LSTM-GARCH and its baselines for test set 3}
\label{LG28}
  \vspace{-20pt}
\end{figure}

\subsubsection{Results of Learning with Statistic Extraction}
We compare the performance  of LSTM-GARCH model with those of ARIMA-GARCH model and LSTM model in  Table \ref{tab2}. shows the prediction results for LSTM,ARIMA-GARCH and LSTM-GARCH for all the test sets. LSTM-GARCH is seen to perform the best for all metrics in all testsets. For the testset 1, compared with the LSTM model, the improvement of MSE, RMSE, MAE and MAPE are 
$61.41\%$,$37.88\%$,$47.48\%$,$47.29\%$ respectively. This demonstrates that the introduction of additional statistic features into the learning platform has a big impact on the performance of LSTM model. Compared with  ARIMA-GARCH, the improvements of MSE, RMSE, MAE and MAPE are 
$20.34\%$,$10.77\%$,$15.85\%$,$15.82\%$, which are still big. 

\begin{table}[]
\footnotesize
\begin{tabular}{|c|c|c|c|c|c|}
\hline
 &  & MSE & RMSE & MAE & MAPE \\ \hline
 & AG & 156.93 & 12.53 & 10.13 & 0.57 \\ \cline{2-6} 
Test Set 1 & LSTM & 323.92 & 18.00 & 16.23 & 0.90 \\ \cline{2-6} 
 & \begin{tabular}[c]{@{}c@{}}LSTM-\\ GARCH\end{tabular} & \textbf{125.01} & \textbf{11.18} & \textbf{8.53} & \textbf{0.48} \\ \hline
 & AG & 155.33 & 12.46 & 10.56 & 0.60 \\ \cline{2-6} 
Test Set 2 & LSTM & 280.57 & 16.75 & 14.38 & 0.81 \\ \cline{2-6} 
 & \begin{tabular}[c]{@{}c@{}}LSTM-\\ GARCH\end{tabular} & \textbf{149.39} & \textbf{12.22} & \textbf{10.05} & \textbf{0.57} \\ \hline
 & AG & 287.66 & 16.96 & 13.92 & 0.78 \\ \cline{2-6} 
Test Set 3 & LSTM & 440.38 & 20.99 & 17.34 & 0.97 \\ \cline{2-6} 
 & \begin{tabular}[c]{@{}c@{}}LSTM-\\ GARCH\end{tabular} & \textbf{285.41} & \textbf{16.89} & \textbf{13.70} & \textbf{0.77} \\ \hline
\end{tabular}
\vspace{-5pt}
\caption{\label{tab2} Comparison of the extraction-based LSTM-GARCH with it's baselines on different test sets}
\vspace{-20pt}
\end{table}




Further, comparing the results in different test sets of table \ref{tab2}, the performance improvement of test set 1 is  the biggest. Over the test set 3,  compared with the LSTM model, the values of MSE, RMSE, MAE and MAPE  improve  $35.2\%$, $19.50\%$, $20.95\%$, and $20.54\%$.  Although the improvements are smaller than those for the test set 1, they are still large. This shows our second method has a  great potential in improving the performance of LSTM model. 




In Figures \ref{LG7} and \ref{LG28},  our proposed method has the best performance and is closest to the curves with true values in all testsets,  and the improvement is more obvious in the first testset.

In Table \ref{tab3}, we further compare the performance between the methods 1 and 2.  ARIMA-GARCH-LSTM not only outperforms the baselines ARIMA-GARCH and LSTM in all metrics, but  is also superior to our proposed  LSTM-GARCH model. This is because that the decomposition-based method could better exploit the features of data to represent different components with appropriate strategies,  while LSTM-GARCH just utilizes the GARCH term to help improve the LSTM performance. However, the second method is more concise and convenient to use, while the first method needs more data pre-processing.

\subsubsection{Robust Analysis}
We use two more data sets to test the superiority of our schemes and the robustness of the performance. The first one is the stock open Price of "Fidelity National Information Services(FIS)" from 01-02-2019 to 12-18-2019 and the second one is the stock open Price of "Walmart(WMT)" from 01-02-2019 to 12-18-2019. 
Table 4 and 5 shows the result of "FIS" and "WMT", respectively.  
For all the test sets, our proposed two schemes perform much better than the ones purely based on model or purely learnt from data.  

\subsubsection{Conclusion}
The increasing demands on big data analysis call for more advanced and accurate learning techniques. Model-based learning and data-based learning are two tools that have been playing important roles. We systematically study the features of data and tools and propose two different methods to concurrently exploit the powerful statistical models and machine learning techniques. We illustrate our scheme using time series data as an example. In the first method, we decompose the time series data into two parts, and use statistical models to represent the stable part and the data-driven learning to track the remaining unstable part. In the second method, we first extract statistic features of data, and then feed them together data as input into the machine learning model. Extensive performance studies demonstrate the superiority of our proposed methods in all experimental settings.  We expect our work will motivate more studies to explore the seamless integration of mathematical models and data-driven schemes to enable reliable and interpretable learning.      

\begin{table}[]
\footnotesize
\begin{tabular}{|l|l|l|l|l|l|}
\hline
 &  & MSE & RMSE & MAE & MAPE \\ \hline
{Test Set 1} & \begin{tabular}[c]{@{}l@{}}AG-\\ LSTM\end{tabular} & 108.32 & 10.41 & 8.02 & 0.45 \\ \cline{2-6} 
\multicolumn{1}{|c|}{} & \begin{tabular}[c]{@{}l@{}}LSTM-\\ GARCH\end{tabular} & 125.01 & 11.18 & 8.53 & 0.48 \\ \hline
{Test Set 2} & \begin{tabular}[c]{@{}l@{}}AG-\\ LSTM\end{tabular} & 108.06 & 10.40 & 8.34 & 0.47 \\ \cline{2-6} 
 & \begin{tabular}[c]{@{}l@{}}LSTM-\\ GARCH\end{tabular} & 149.39 & 12.22 & 10.05 & 0.57 \\ \hline
{Test Set 3} & \begin{tabular}[c]{@{}l@{}}AG-\\ LSTM\end{tabular} & 263.40 & 16.23 & 12.35 & 0.70 \\ \cline{2-6} 
 & \begin{tabular}[c]{@{}l@{}}LSTM-\\ GARCH\end{tabular} & 285.41 & 16.89 & 13.70 & 0.77 \\ \hline
\end{tabular}
\vspace{-5pt}
\caption{\label{tab3} Comparison between AG-LSTM and LSTM-GARCH for all test sets.}
\vspace{-5pt}
\end{table}

\begin{table}[]
\vspace{-10pt}
\footnotesize
\begin{tabular}{|l|l|l|l|l|l|}
\hline
 &  & MSE & RMSE & MAE & MAPE \\ \hline
 & AG & 2.33 & 1.53 & 1.30 & 0.98 \\ \cline{2-6} 
 & LSTM & 2.29 & 1.52 & 1.20 & 0.90 \\ \cline{2-6} 
{Test Set 1} & \begin{tabular}[c]{@{}l@{}}LSTM-\\ GARCH\end{tabular} & 2.03 & 1.43 & 1.17 & 0.88 \\ \cline{2-6} 
\multicolumn{1}{|c|}{} & \begin{tabular}[c]{@{}l@{}}AG-\\ LSTM\end{tabular} & 1.70 & 1.30 & 1.18 & 0.89 \\ \hline
& AG & 1.95 & 1.40 & 1.16 & 0.87 \\ \cline{2-6} 
 & LSTM & 2.55 & 1.60 & 1.27 & 0.95 \\ \cline{2-6} 
{Test Set 2} & \begin{tabular}[c]{@{}l@{}}LSTM-\\ GARCH\end{tabular} & 1.93 & 1.39 & 1.06 & 0.79 \\ \cline{2-6} 
 & \begin{tabular}[c]{@{}l@{}}AG-\\ LSTM\end{tabular} & 1.54 & 1.24 & 1.04 & 0.77 \\ \hline
 & AG & 2.11 & 1.45 & 1.13 & 0.84 \\ \cline{2-6} 
 & LSTM & 2.41 & 1.55 & 1.28 & 0.95 \\ \cline{2-6} 
{Test Set 3} & \begin{tabular}[c]{@{}l@{}}LSTM-\\ GARCH\end{tabular} & 2.10 & 1.45 & 1.05 & 0.78 \\ \cline{2-6} 
 & \begin{tabular}[c]{@{}l@{}}AG-\\ LSTM\end{tabular} & 1.93 & 1.38 & 1.11 & 0.82 \\ \hline
\end{tabular}
\vspace{-5pt}
\caption{\label{tab4} Comparison of different schemes over data set 2}
\vspace{-15pt}
\end{table}

\begin{table}[]
\footnotesize
\begin{tabular}{|c|c|c|c|c|c|}
\hline
 &  & MSE & RMSE & MAE & MAPE \\ \hline
 & AG & 2.82e-4 & 1.68e-2 & 1.37e-2 & 1.08 \\ \cline{2-6} 
 & LSTM &  0.84e-4 & 2.90e-2 & 2.47e-2 & 1.89 \\ \cline{2-6} 
{Test 1} & \begin{tabular}[c]{@{}l@{}}LSTM-\\ GARCH\end{tabular} & 1.99e-4 & 1.41e-2 & 1.20e-2 & 0.94 \\ \cline{2-6} 
\multicolumn{1}{|c|}{} & \begin{tabular}[c]{@{}l@{}}AG-\\ LSTM\end{tabular} & 1.60e-4 & 1.27e-2 & 1.13e-2 & 0.89 \\ \hline
& AG & 2.32e-4 & 1.52e-2 & 1.27e-2 & 1.01 \\ \cline{2-6} 
 & LSTM & 8.71e-4 & 2.95e-2 & 2.68e-2 & 2.11 \\ \cline{2-6} 
{Test 2} & \begin{tabular}[c]{@{}l@{}}LSTM-\\ GARCH\end{tabular} & 1.77e-4 & 1.33e-2 & 1.13e-2 & 0.91 \\ \cline{2-6} 
 & \begin{tabular}[c]{@{}l@{}}AG-\\ LSTM\end{tabular} & 1.69e-4 & 1.30e-2 & 1.15e-2 & 0.92 \\ \hline
 & AG & 4.82e-4 & 2.19e-2 & 1.67e-2 & 1.39 \\ \cline{2-6} 
 & LSTM & 9.20e-4 & 3.03e-2 & 2.75e-2 & 2.22 \\ \cline{2-6} 
{Test 3} & \begin{tabular}[c]{@{}l@{}}LSTM-\\ GARCH\end{tabular} & 4.43e-4 & 2.11e-2 & 1.59e-2 & 1.33 \\ \cline{2-6} 
 & \begin{tabular}[c]{@{}l@{}}AG-\\ LSTM\end{tabular} & 4.19e-4 & 2.05e-2 & 1.57e-2 & 1.31 \\ \hline
\end{tabular}
\vspace{-5pt}
\caption{\label{tab5} Comparison of different schemes over data set 3}
\end{table}

\bibliography{aaai21.bib} 

\end{document}